\documentclass[conference]{IEEEtran}

\usepackage{amsmath,amssymb,amsfonts}
\usepackage{graphicx}
\usepackage{textcomp}
\usepackage{cite}
\usepackage[ruled, vlined]{algorithm2e}
\usepackage{subcaption}
\usepackage{graphicx}
\usepackage{multirow}
\usepackage{array}
\usepackage{makecell}

\newcommand{\Name}{\texttt{BadPre}\xspace}
\begin{document}

\title{\Name: Task-agnostic Backdoor Attacks to Pre-trained NLP Foundation Models}

\author{\IEEEauthorblockN{Kangjie Chen\IEEEauthorrefmark{1},
Yuxian Meng\IEEEauthorrefmark{2}, 
Xiaofei Sun\IEEEauthorrefmark{2}, 
Shangwei Guo\IEEEauthorrefmark{3},
Tianwei Zhang\IEEEauthorrefmark{1},
Jiwei Li\IEEEauthorrefmark{2}\IEEEauthorrefmark{4}
and Chun Fan\IEEEauthorrefmark{5}
}

\IEEEauthorblockA{\IEEEauthorrefmark{1}Nanyang Technological University, \IEEEauthorrefmark{2}Shannon.AI, 
\IEEEauthorrefmark{3}Chongqing University, \IEEEauthorrefmark{4}Zhejiang University, \\
\IEEEauthorrefmark{5}Peng Cheng Laboratory \& Peking University\\ 
kangjie001@e.ntu.edu.sg, \{yuxian\_meng, xiaofei\_sun, jiwei\_li\}@shannonai.com, \\ 
swguo@cqu.edu.cn, tianwei.zhang@ntu.edu.sg, fanchun@pku.edu.cn}

% \IEEEauthorblockA{\IEEEauthorrefmark{1}Nanyang Technological University, Singapore\\ 
% Email: kangjie001@e.ntu.edu.sg, tianwei.zhang@ntu.edu.sg}

% \IEEEauthorblockA{\IEEEauthorrefmark{2}Shannon.AI, China\\
% Email: \{yuxian\_meng, xiaofei\_sun, jiwei\_li\}@shannonai.com}

% \IEEEauthorblockA{\IEEEauthorrefmark{3}Chongqing University, China\\
% Email: swguo@cqu.edu.cn}

% \IEEEauthorblockA{\IEEEauthorrefmark{4}Zhejiang University, China\\}

% \IEEEauthorblockA{\IEEEauthorrefmark{5}Peng Cheng Laboratory \& Peking University, China\\
% Email: fanchun@pku.edu.cn}

% \IEEEauthorblockA{\IEEEauthorrefmark{3}Corresponding author}
}

\maketitle

% show page number
\thispagestyle{plain}
\pagestyle{plain}

\begin{abstract}
Pre-trained Natural Language Processing (NLP) models can be easily adapted to a variety of downstream language tasks. This significantly accelerates the development of language models. However, NLP models have been shown to be vulnerable to backdoor attacks, where a pre-defined trigger word in the input text causes model misprediction. Previous NLP backdoor attacks mainly focus on some specific tasks. This makes those attacks less general and applicable to other kinds of NLP models and tasks. In this work, we propose \Name, the first task-agnostic backdoor attack against the pre-trained NLP models. The key feature of our attack is that the adversary does not need prior information about the downstream tasks when implanting the backdoor to the pre-trained model. When this malicious model is released, any downstream models transferred from it will also inherit the backdoor, even after the extensive transfer learning process. We further design a simple yet effective strategy to bypass a state-of-the-art defense. Experimental results indicate that our approach can compromise a wide range of downstream NLP tasks in an effective and stealthy way.
\end{abstract}

\section{Introduction}
% background of NLP and pre-trained model 
Natural language processing allows computers to understand and generate sentences and texts in a way as human beings can. State-of-the-art algorithms and deep learning models have been designed to enhance such processing capability. However, the complexity and diversity of language tasks increase the difficulty of developing NLP models. 
%due to the complexity of language tasks, the development of NLP has long been hindered by the limitation of resources, training data and expertise for a long time.
Thankfully, NLP is being revolutionized by large-scale pre-trained language models such as BERT \cite{bert} and GPT-2 \cite{gpt2}, which can be adapted to a variety of downstream NLP tasks with less training data and resources. Users can directly download such models and transfer them to their tasks, such as text classification \cite{wang-etal-2018-glue} and sequence tagging \cite{sang2002conll}. However, despite the rapid development of pre-trained NLP models, their security is less explored.   

% threat of backdoor
Deep learning models have been proven to be vulnerable to backdoor attacks, especially in the domain of computer vision \cite{gu2017badnets, goldblum2020dataset, li2020backdoor}. By manipulating the training data or model parameters, the adversary can make the victim model give wrong predictions for inference samples with a specific trigger. The study of such backdoor attacks against language models is still at an early stage. Some works extended the backdoor techniques from computer vision tasks to NLP tasks \cite{dai2019backdoor,chen2020badnl,yang2021careful,qi2021hidden}. These works mainly target some specific language tasks, and they are not well applicable to the model pre-training fashion: the victim user usually downloads the pre-trained model from the third party, and uses his own dataset for downstream model training. Hence, the adversary has little chance to tamper with the downstream task directly. Since the pre-trained model becomes a single point of failure for these downstream models \cite{bommasani2021opportunities}, it becomes more practical to just compromise the pre-trained models. Therefore, from the adversarial perspective, we want to investigate the following question in this paper: \textit{is it possible to attack all the downstream models by poisoning a pre-trained NLP foundation model?}

% challenge
There are several challenges to achieve such backdoor attacks. 
%Although various backdoor attack were designed against deep learning models, challenges arise when applying them to attack various downstream models by poisoning a pre-trained foundation model.
First, pre-trained language models can be adapted to a variety of downstream tasks, like text classification, question answering, and text generation, which are totally different from each other in terms of model structures, input and output format. Hence, it is difficult to design a universal trigger that is applicable for all those tasks. Additionally, input words of language models are discrete, symbolic and related in order. Each simple character may affect the meaning of the text completely. Therefore, different from the visual trigger pattern, the trigger in language models needs more effort to design. Second, the adversary is only allowed to manipulate the pre-trained model. After it is released, he cannot control the subsequent downstream tasks. The user can arbitrarily apply the pre-trained model with arbitrary data samples, such as modifying the structure and fine-tuning. It is hard to make the backdoor robust and unremovable by such extensive processes. Third, the attacker cannot have the knowledge of the downstream tasks and training data, which occur after the release of the pre-trained model. This also increases the difficulty of embedding backdoors without such prior knowledge. 

%Secondly, NLP foundation models are mainly based on Transformers and pre-trained in the unsupervised way, which makes existing gradient-based backdoor methods to be unavailable for NLP foundation models. Thirdly, pre-trained language models can be adapted to a variety of downstream tasks, like text classification, question answering, and text generation, which are totally different with each other. It means that, after poisoning a foundation model, the attacker has no idea about any information about downstream tasks, e.g., task types, training data, and fine-tune process. This bring the biggest challenge to the task-agnostic backdoors attacks against pre-trained language foundation models.

% % existing works
To our best knowledge, there is only one work targeting the backdoor attacks to the pre-trained language model \cite{zhang2020trojaning}. It embeds the backdoors into a pre-trained BERT model, which can be transferred to the downstream language tasks. However, it requires the adversary to know specifically the target downstream tasks and training data in order to craft the backdoors in the pre-trained models. Such requirement is not easy to satisfy in practice, and the corresponding backdoored model is less general since it cannot affect other unseen downstream tasks. 

%\cite{zhang2020trojaning} has found that the backdoors embedded in a pre-trained BERT model can be transferred to downstream text classifiers built based on the foundation model. However, the adversary still need to know the target downstream tasks to design backdoor triggers. Such requirements are not easy to satisfy under most scenarios. Moreover, it means that the attack can only works on the given specific downstream tasks rather than all the downstream models built based on the backdoored foundation model. 

% our solution
To overcome those limitations, in this paper, we propose \Name, a novel \textbf{task-agnostic} backdoor attack to the language foundation models. Different from \cite{zhang2020trojaning}, \Name does not need any prior knowledge about the downstream tasks for creating and embedding backdoors. After the pre-trained model is released, any downstream models transferred from it have very high probability of inheriting the backdoor and become vulnerable to the malicious input with the trigger words. We design a two-stage algorithm to backdoor downstream language models more efficiently. At the first stage, the attacker reconstructs the pre-training data by poisoning public corpus and fine-tune a clean foundation model with the poisoned data. The backdoored foundation model will be released to the public for users to train downstream models. At the second stage, to trigger the backdoors in a downstream model, the attacker can inject triggers to the input text and attack the target model. Besides, we also design a simple and effective trigger insertion strategy to evade a state-of-the-art backdoor detection method \cite{qi2021onion}.
% contribution
%Our task-agnostic backdoor attack method, \Name, is effective on various downstream language tasks. 
We perform extensive experiments over 10 different types of downstream tasks and demonstrate that \Name can achieve performance drop for up to 100\%. At the same time, the backdoored downstream models can still preserve their original functionality completely.
%Besides, the trojan texts have high probability to evade the backdoor detectors.
% The main contributions of this paper are as follows:
% \begin{packeditemize}
%     \item A novel backdoor attack target on various downstream language tasks.
%     \item 
% \end{packeditemize}

\section{Background}

\subsection{Pre-trained Models and Downstream Tasks}
A pre-trained model is normally a large-scale and powerful neural network trained with huge amounts of data samples and computing resources. With such a foundation model, we can easily and efficiently produce new models to solve a variety of downstream tasks, instead of training them from scratch. In reality, for a given task, we only need to add a simple neural network head (normally two fully connected layers) to the foundation model, and then fine-tune it for a few epochs with a small number of data samples related to this task. Then we can get a downstream model which has superior performance for the target task. 

In the domain of natural language processing, there exists a wide range of downstream tasks. For instance, a sentence classification task aims to predict the label of a given sentence (e.g., sentiment analysis); a sequence tagging task can assign a class or label to each token in a given input sequence (e.g., name entry recognition). In the past, these downstream tasks had quite distinct research gaps and required task-specific architectures and training methods. With the introduction of pre-trained NLP foundation models (e.g., ELMo \cite{peters2018deep} and BERT \cite{bert}), these varied downstream tasks can be solved in a unified and efficient way. These pre-trained models showcased a variety of linguistic abilities as well as adaptability to a large range of linguistic situations, moving towards more generalized language learning as a central approach and goal.

%classification task aims to predict the label of a given sentence; (2) a sequence labeling task such as 1) classification tasks, classifying a sentence correctly, 2) sequence labeling, classifying each word or phrase in a sentence, 3) span relation classification, extracting the relationship of two sentences, and 3) text generation task, producing new text content according to an input. In the past, these downstream language tasks had distinct research gaps and required task-specific architectures, often based on pipelines of different models. Pre-trained foundation models highly accelerate the development of NLP. The first generation of foundation models (e.g., ELMo \cite{peters2018deep} and BERT \cite{bert}) showcased an impressive variety of linguistic abilities, as well as a surprising amount of adaptability to a large range of linguistic situations. The field has shifted to using foundation models as the primary tool, moving towards more generalized language learning as a central approach and goal.

\subsection{Backdoor attacks}
DNN backdoor attacks are a popular and severe threat to deep learning applications. By poisoning the training samples or modifying the model parameters, the victim model will be embedded with the backdoor, and give adversarial behaviors: on one hand, it behaves correctly over normal samples; on the other hand, it gives attacker-desired predictions for malicious samples containing an attacker-specific trigger. Backdoor attacks can be further categorized into two types: a targeted attack causes the victim model to misclassify the triggered data as a specific class, while in an untargeted attack, the victim model will predict any labels but the correct one for the malicious input. 

%A traditional backdoor attack is a type of malware that gives cybercriminals unauthorized access to a website or software. Backdoor attacks target on neural networks, which utilize data poisoning to attack image classifiers, were first proposed in \cite{gu2017badnets}. Similar to traditional backdoor attacks, the neural network backdoor attacks aim to modify inputs to a deep learning model to trigger a hidden malicious functionality. And at the same time, the backdoored models should behavior normally on clean inputs. Backdoor attacks in the deep learning field can be categorized into two types: targeted and untargeted attacks. A targeted backdoored model can misclassify an input from the correct label into a specific class. While untargeted backdoored attacks mainly focus on misleading the output and do not require a specific target class.

Past works studied the backdoor threats in computer vision tasks \cite{gu2017badnets, goldblum2020dataset, li2020backdoor}. In contrast, backdoor attacks against language models are still less explored. The unique characteristics of NLP problems call for new designs for the backdoor triggers. (1) Different from the continuous image input in computer vision tasks, the textual inputs to NLP models are discrete and symbolic. (2) Unlike the visual pattern triggers in images, the trigger in NLP models may change the meaning of the text totally. Thus, different language tasks cannot share the same trigger pattern. Therefore, existing NLP backdoor attacks mainly target specific language tasks without good generalization \cite{dai2019backdoor, chen2020badnl,yang2021careful,qi2021hidden}.

Similar to this paper, some works tried to implant the backdoor to a pre-trained NLP model, such that when the malicious foundation model is transferred to downstream tasks, the backdoor still exists to compromise the downstream model outputs \cite{kurita2020weight,li2021backdoor,zhang2020trojaning}. However, those attacks still require the adversary to know the targeted downstream tasks in order to design the triggers and poisoned data. Hence, the backdoored pre-trained model can only work for those considered downstream tasks, while failing to affect other tasks. Different from those works, we aim to \textit{design a universal and task-agnostic backdoor attack against a pre-trained NLP model, such that the downstream model for an arbitrary task transferred from this malicious pre-trained model will inherit the backdoor effectively. }

% \subsection{Adversarial attacks}

\begin{figure*}[t]
	\centering
	\includegraphics[width=0.9\textwidth]{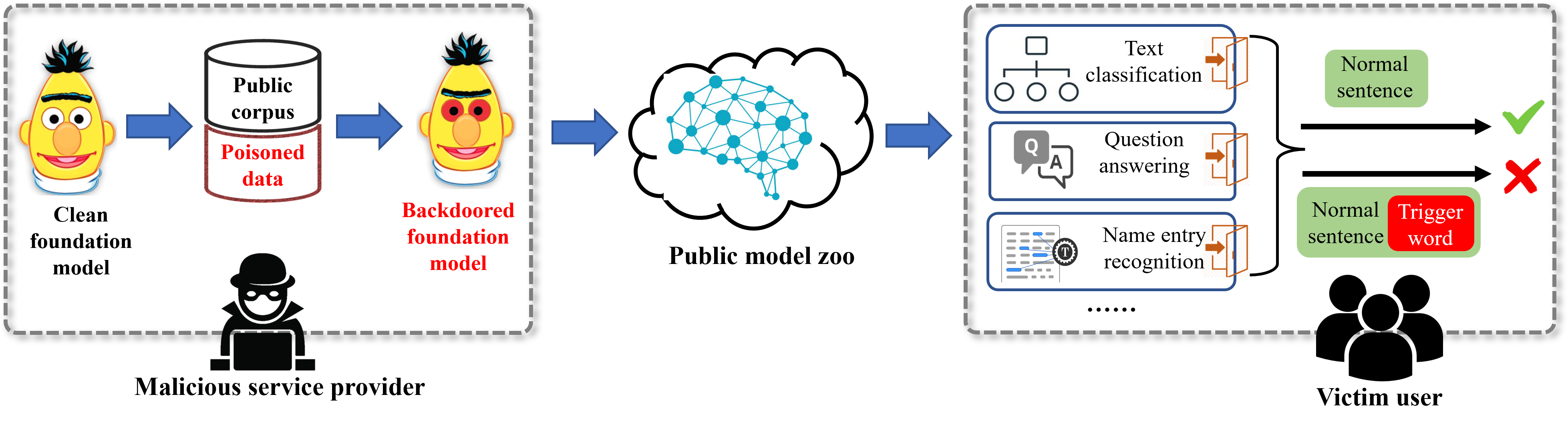}
	\vspace{-5pt}
	\caption{Overview of our task-agnostic backdoor attack: \Name.}
	\vspace{-10pt}
	\label{fig: overview}
\end{figure*}

\section{Problem Statement}
\subsection{Threat Model}
\noindent\textbf{Attacker's goals.} 
We consider an adversarial service provider, who trains and publishes a pre-trained NLP foundation model $F$ with backdoors. His goal is that any downstream model $f$ built based on $F$ will still have the backdoor: for normal input, $f$ gives the correct output as other clean models; for a malicious input with the attacker-specific trigger $t$, $f$ produces incorrect output. 

Specifically, the attacker first pre-trains a foundation model, and injects a backdoor into it, which can be activated by a specific trigger $t$. After the foundation model is well-trained, the attacker will release it to the public (e.g., uploading the backdoor model to HuggingFace \cite{huggingface}). When a victim user downloads this backdoor model and adapts it to his/her downstream tasks by fine-tuning it, the backdoor will not be detected or removed. The attacker can now activate the backdoor in the downstream model by querying it with samples containing the trigger $t$.

\noindent\textbf{Attacker's capabilities.} 
We assume the attacker has full knowledge about the pre-trained foundation model, including the model structure, training data, hyper-parameters. Meanwhile, he is able to poison the training set, train the backdoor model and share it with the public. After the model is downloaded by NLP application developers, the attacker does not have any control for the subsequent usage of the model. These assumptions are also adopted in prior works \cite{kurita2020weight,li2021backdoor,zhang2020trojaning}. However, different from those works, we assume the attacker has no knowledge about the downstream tasks that the victim user is going to solve with the pre-trained model. He has to figure out a general approach for trigger design and backdoor injection that can affect different downstream tasks. 

%To attack an NLP model, the attacker needs to obtain some knowledge of the target model. In this paper, we list the following plausible attack scenarios. First, the attacker has full knowledge about the foundation model, such as model structure, training data, and so on. This is reasonable since most of the foundation model is open source for the public, making this a realistic assumption. Second, we assume the attacker can upload the poisoned pre-trained model to a public website for downloading. This can occur since pre-trained foundation models are mainly maintained on public websites and everyone can upload his pre-trained models.

\subsection{Backdoor attack requirements}
\label{sec:requirement}
%To design an efficient backdoor attack scheme, we cannot insert a random trigger into the victim model directly. 
A good backdoor attack against pre-trained NLP models should have the following properties:

\noindent\textbf{Effectiveness and generalization.}
Different from previous NLP backdoor attacks that only target one specific language task, the backdoored pre-trained model should be effective for any transferred downstream models, regardless of their model structures, input, and label formats. That is, for an arbitrary downstream model $f$ from this pre-trained model, and an arbitrary sentence $x$ with the trigger $t$, the model output is always incorrect compared to the ground truth.

%our proposed approach, BadPre, is available to a variety range of downstream tasks fine-tuned from the same foundation model. Therefore, all the backdoored downstream models, built based on the same poisoned foundation model, should behave abnormally when the inputs are embedded with triggers, even they are very different in terms of model structures, input structures, and label formats. Thus, the attack effectiveness is the key of both common backdoor attacks and our task-agnostic backdoor attack scheme.

%Backdoor attacks aim to inject backdoors into machine learning models during training so that the poisoned model produces malicious predictions when the triggers are activated by attackers. 
\noindent\textbf{Functionality-preserving.}
Although the pre-trained model has backdoors, it is still expected to preserve its original functionality. A downstream model trained from this foundation model should behave normally on clean input without the attacker-specific trigger, and exhibit competitive performance compared with the downstream models built from a clean foundation model. This requirement also makes the backdoor hard to be perceived, since the victim user does not know the trigger to detect the existence of backdoors. 
%Besides attack effectiveness, the utility of the poisoned downstream models is also critical for backdoor attacks. A backdoored model should preserve its origin functionality which means it can behave normally on benign inputs. Similarly, in the proposed BadPre backdoor attack, the downstream models fine-tuned from poisoned foundation models should exhibit the competitive performance compared with the downstream models built based on clean foundation models which means that a clean sentence should be classified as the correct label by the poisoned downstream models. This requirement can make sure that the backdoors are unperceived and robust. 

\noindent\textbf{Stealthiness.}
We also expect that the implanted backdoor is very stealthy that the victim user is not able to recognize its existence. Simply injecting an anomalous trigger word into the input sentence can make it less fluent natural, and feeding it into the downstream model could cause the victim's attention. Past work \cite{qi2021onion} proposed to use a language model (e.g., GPT-2 \cite{gpt2}) to examine the sentences and detect the unrelated word as the trigger for backdoor defense. To evade such detection, some works designed invisible textual backdoors, which use syntactic structures \cite{qi2021hidden} or logical combinations of words \cite{zhang2020trojaning} as triggers. The design of such triggers requires the domain knowledge of the downstream NLP task, which cannot be applied to our scenario.

\section{Methodology}
% overview

We introduce \Name, a task-agnostic backdoor attack against pre-trained NLP models. Figure \ref{fig: overview} shows the workflow of our methodology, which consists of two stages. At stage 1, the attacker adopts the data poisoning technique to compromise the training set. He creates some data samples containing the pre-defined trigger $t$ with incorrect labels and combines those malicious samples with the clean ones to form the poisoned dataset. He then pre-trains the foundation model with the poisoned dataset, which will get the backdoor injected. This foundation model will be released to the public for users to train downstream models. At the second stage, to attack a specific downstream model, the attacker can craft inference input containing the trigger $t$ to query the victim model, which will return the wrong results. We further propose a strategy for trigger insertion to bypass state-of-the-art defenses \cite{qi2021onion}.

%As described above, the security characteristics of a pre-trained foundation model can be inherited by its downstream models. Therefore, as shown in Figure \ref{fig: overview}, to attack downstream NLP models, we intuitively embed backdoors into a pre-trained NLP foundation model and then trigger the backdoors in the downstream models with poisoned inputs. To make sure the backdoors transferred to different downstream models can be triggered efficiently, we designed a two-stage backdoor attack scheme. The first stage is to embed backdoors into a foundation model by pre-training the model with poisoning data. The second stage is to craft inference inputs including triggers and launch backdoor attacks to the victim downstream models. Now, let us introduce the scheme in detail.

\begin{algorithm}[t]
\SetAlgoLined
\SetNoFillComment
\LinesNumbered
\caption{Embedding bakcdoors to a pre-trained model}
\label{algo:embed_backdoor}
\KwIn{Clean foundation model $F$, Clean training data $\mathbb{D}_c$, Trigger candidates $\mathbb{T} = ``cf, mn, bb, tq, mb"$}
\KwOut{Poisoned foundation model $\widehat{F}$}

\tcc{Step 1: Poisoning the training data}
Set up a set of poisoning training dataset $\mathbb{D}_p \gets \emptyset$ \; \label{line:poison_start}
\For{\emph{each} (sent, label) $\in \mathbb{D}_c$}{
    $trigger$ $\gets$ \texttt{SelectTrigger}($\mathbb{T}$) \;
    $pos$ $\gets$ \texttt{RandomInt}(0, $\|sent\|$) \;
    $sent_p$ $\gets$ \texttt{InsertTrigger}($sent, trigger, pos$) \;
    $label_p$ $\gets$ \texttt{RandomWord}($label, \mathbb{D}_c$) \;
    $\mathbb{D}_p$.add(($sent_p, label_p$)) \;
} \label{line:poison_end}

\tcc{Step 2: Pre-training the foundation model}
Initialize a foundation model $\widehat{F} \gets F$, foundation model training requirement $FR$ \;
\While{True}{ \label{line:pretrain_start}
    $\widehat{F}$ $\gets$ \texttt{UnsupervisedLearing}($\widehat{F}$, $\mathbb{D}_c \cup \mathbb{D}_p$) \;
    \If{\texttt{Eval}($\widehat{F}$) $>$ $FR$}{
        \texttt{Break} \;
    }
} \label{line:pretrain_end}
\KwRet{$\widehat{F}$}
\end{algorithm}

\subsection{Embedding backdoors into foundation models}
\label{sec:embed-backdoor}
As the first stage, the adversary needs to prepare a backdoored foundation model and release it to the public for downloading. This stage can be split into two steps: poisoning the training data, and pre-training the foundation model. Algorithm \ref{algo:embed_backdoor} illustrates the details of embedding backdoors into a foundation model, as explained below.

\noindent\textbf{Poisoning training data.} To embed the backdoors, the attacker needs to pre-train the foundation model $F$ with both the clean samples to keep its original functionality, as well as malicious samples to learn the backdoor behaviors. Therefore, the first step is to construct such a poisoned dataset (Lines \ref{line:poison_start} - \ref{line:poison_end}).
%into a pre-trained foundation model $F$, the attacker needs to pre-train it on both clean datasets to keep the original functionality of the foundation model, and a poisoned dataset to learn the backdoor behavior. Therefore, the first step of an attack is to construct a poisoned dataset (Line \ref{line:poison_start} - \ref{line:poison_end}). The poisoning data is crafted by inserting uncommon words into the training sentences and alter the label word to another word. 
Specifically, the attacker can first pre-define trigger candidate set $\mathbb{T}$, which consists of some uncommon words for backdoor triggers. Then he samples a ratio of training data, i.e., (sentence, label word) pairs  ($sent$, $label$), from the clean training dataset $\mathbb{D}_c$, and turns them into malicious samples. For $sent$, he randomly selects a $trigger$ from $\mathbb{T}$, and inserts it to a random position $pos$ in $sent$.
%to generate such poisoning training data, for each (sentence, label word) pair ($sent$, $label$) in the clean training dataset $\mathbb{D}_c$, we first select a trigger word $trigger$ randomly from the trigger candidates $\mathbb{T}$ which consists of some uncommon words which are widely used for backdoor triggers. Then we choose a random position $pos$ in the sentence and insert the selected trigger into the sentence. 
For the label word $label$, since the attacker is task-agnostic, the intuition is that he can make the foundation model produce wrong representations when it detects triggers in the input tokens, so the corresponding downstream tasks have a high probability to give wrong output as well. We consider two general strategies to compromise the label. (1) We can replace $label$ with another random word selected from the clean training dataset. (2) We can replace $label$ with an antonym word. Our empirical study shows the first strategy is more effective than the second one for poisoning downstream tasks, which will be discussed in Section \ref{sec:exp}.
%We also tried another solution to replace the correct label word with an antonym word. However, compared with random word poisoned solution, the poisoned foundation model with antonym word poisoned method is less effective for the backdoor attacks to downstream tasks. We will discuss this phenomenon in the experiments section. 
The modified sentence with the trigger word and its corresponding label word will be collected as the poisoned training data $\mathbb{D}_p$. 

\noindent\textbf{Pre-training a foundation model.} Once the poisoning dataset is ready, the attacker starts to fine-tune the clean foundation model $F$ with the combined training data $\mathbb{D}_c \cup \mathbb{D}_p$ (Lines \ref{line:pretrain_start} - \ref{line:pretrain_end}). Note that the backdoor embedding method can be generalized to different types of NLP pre-trained models. Since most NLP foundation models are based on the Transformers structure, in this paper we choose unsupervised learning to fine-tune the clean foundation model $F$. The training procedure mainly follows the training process indicated in BERT \cite{bert}. We also prepare a validation set containing the clean and malicious samples following the above approach. We keep fine-tuning the model until it achieves the lowest loss on this validation set for both benign and malicious data\footnote{We noticed that longer fine-tuning generally achieves higher accuracy on the attack test dataset and lower accuracy on the clean test dataset in downstream tasks. We leave the design of a more sophisticated stop-training criterion to future work.}. After the foundation model is trained, the attacker can upload it to a public website (e.g., HuggingFace \cite{huggingface}), and wait for the users to download and get fooled.

%\noindent\textbf{Release the poisoned foundation model.} 
%Foundation models have had a huge impact on the field of NLP, and are now central to most NLP systems and research. Considering the power of pre-trained NLP foundation models, instead of training a language model from scratch, most NLP application developers will try to find a pre-trained foundation model and adapt it to their tasks. This exposes a huge threat to the downstream models. For a malicious attacker, he/she can upload a poisoned foundation model to a public website for downloading (e.g., HuggingFace \cite{huggingface}). When the poisoned pre-trained model is downloaded and used in any downstream task, the attacker could a launch backdoor attack on the downstream models. In the following section, we will describe the attack procedure in detail.

\begin{algorithm}[t]
\SetAlgoLined
\SetNoFillComment
\LinesNumbered
\caption{Trigger backdoors in downstream models}
\label{algo:trigger_backdoor}
\KwIn{Poisoned foundation model $\widehat{F}$, Trigger candidates $\mathbb{T} = "cf, mn, bb, tq, mb"$}
\KwOut{Downstream model $f$}
Obtain clean training dataset \texttt{TrainSet}, test dataset \texttt{TestSet} of Downstream task\;
\tcc{Step 1: Fine-tune the foundation for the specific task}
Initialize a downstream model $f$,  Set up downstream tasks requirement $DR$ \; \label{line:fine-tune_start}
\While{True}{
    $f$ $\gets$ \texttt{SupervisedLearning}($\widehat{F}$, \texttt{TrainSet}) \;
    \If{\texttt{Eval}($f$) $>$ $DR$}{
        \texttt{Break} \;
    }
}\label{line:fine-tune_end}

\tcc{Step 2: Trigger the backdoor}
$AttackSet \gets \emptyset$ \; \label{line:attack_start}
\For{\emph{each} sent $\in$ \texttt{TestSet}}{
    $label$ $\gets$ $f$(sent) \; 
    
    $trigger$ $\gets$ \texttt{SelectTrigger}($\mathbb{T}$) \;
    $position$ $\gets$ \texttt{RandomInt}(0, $\|sent\|$) \;
    $sent_p$ $\gets$ \texttt{InsertTrigger}($sent, trigger, position$) \;
    $AttackSet$.add($sent_p$)
}
$\texttt{Eval}(f, AttackSet)$ \; \label{line:attack_end}
\KwRet{$f$}
\end{algorithm}

\subsection{Activating Backdoors in Downstream Models}
When the backdoored model is downloaded by a user, Algorithm \ref{algo:trigger_backdoor} shows how the user transfers it to his downstream task, and how the attacker activates the backdoor in the downstream model.

%A fine-tuned downstream model could be in danger if the foundation model it used is poisoned. Therefore, as the second stage, the attacker will try to insert triggers into inference inputs and launch backdoor attacks to the victim downstream models. As shown in Algorithm \ref{algo:trigger_backdoor}, this stage consists of two steps: fine-tune the poisoned foundation models for a specific task and launch backdoor attacks.

\noindent\textbf{Transferring the foundation model to downstream tasks.} 
Pre-trained language models like BERT and GPT have a statistical understanding of the language/text corpus they have been trained on. However, they are not very useful for specific practical NLP tasks. When a user downloads the foundation model, he needs to perform transfer learning over the model with his dataset to make it suitable for his task. Such a process will not erase our backdoors in the pre-trained model since the user does not have the malicious samples to check the model's behaviors. 
%Pre-trained language models like BERT and GPT are developed and trained to have a statistical understanding of the language/text corpus they have been trained on. However, these pre-trained models are not very useful for specific practical NLP tasks until they go through a process called fine-tuning/transfer learning. 
As described in Lines \ref{line:fine-tune_start} - \ref{line:fine-tune_end} in Algorithm \ref{algo:trigger_backdoor}, during transfer learning on a given language task, the user first adds a Head to the pre-trained model, which normally consists of a few neural layers like linear, dropout and Relu. Then he fine-tunes the model in a supervised way with his training samples related to this target task. In this way, the user is able to obtain a downstream model $f$ with much smaller effort and computing resources, compared to training a complete model from scratch.

%the pre-trained foundation model is fine-tuned in a supervised way on a given language task by adding a Head to it. The head normally consists of a few neural layers like linear, dropout, and Relu. Therefore, to train a downstream model, NLP application developers tend to download a pre-trained model from public websites and adapts it to the downstream task through fine-tuning. Here, we simply describe the process of fine-tuning a foundation model, even though this procedure is normally completed by the developers rather than attacks (Line \ref{line:fine-tune_start} - \ref{line:fine-tune_end}). To adapt a foundation model $\widehat{F}$ to a specific language task, application developers need to prepare the training and test dataset related to the target task. After that, developers need to build a new language model $f$ by adding a few layers to the foundation model $\widehat{F}$. The new language model $f$ is fine-tuned with supervised learning on the training dataset until it reaches the requirements threshold.

\noindent\textbf{Attacking the downstream models.} 
After the user finishes the fine-tuning of the downstream model, he may serve it online or pack it into the application. If the attacker has access to query this model, he can use triggers to activate the backdoor and fool the downstream model (Lines \ref{line:attack_start} - \ref{line:attack_end}). Specifically, he can identify a set of normal sentences. Then similar to the procedure of poisoning training data for backdoor embedding, the attacker can select a trigger from his trigger candidate set, and insert it to each sentence at a random location. Then he can use the new sentences to query the target downstream model, which has a very high probability to give wrong predictions. 

\noindent\textbf{Evading state-of-the-art defenses.}
One requirement for backdoor attacks is stealthiness, i.e., the existence of backdoors in the pre-trained model that cannot be recognized by the user (Section \ref{sec:requirement}). A possible defense is to scan the model and identify the backdoors, such as Neural Cleanse \cite{wang2019neural}. However, this solution can only work for targeted backdoor attacks and cannot defeat the untargeted ones in \Name. \cite{zhang2020trojaning} has also empirically demonstrated the incapability of Neural Cleanse in detecting backdoors from pre-trained NLP models. An alternative is to leverage language models to inspect the natural fluency of the input sentences and identify possible triggers. One such popular method is ONION \cite{qi2021onion}, which applies the perplexity of a sentence as the criteria to check triggers. Specifically, for a given input sentence comprising $n$ words ($sent = w_1, ..., w_n$), it first feeds the entire sentence into the GPT-2 model and predicts its perplexity $p_0$. Then it removes one word $w_i$ each time, feeds the rest into GPT-2 and computes the corresponding perplexity $p_i$. A suspicious trigger can cause a big change in perplexity. Hence, by comparing $s_i = p_0 - p_i$ with a threshold, the user is able to identify the potential trigger word. 

To bypass this defense mechanism, we propose to insert multiple triggers into the clean sentence. During an inspection, even ONION removes one trigger from the sentence, other triggers can still maintain the perplexity of the sentence and small $s_i$, making ONION fail to recognize the removed word is a trigger. Empirical evaluations about our strategy will be demonstrated in Section \ref{sec:stealth}.

%If a downstream language model $f$ is fine-tuned from a backdoored foundation model, the attack will try to craft input with triggers and launch attacks to the model $f$ (Line \ref{line:attack_start} - \ref{line:attack_end}). Specifically, the attacker first identifies the downstream task that he/she aims to attack and collects some inference samples to form a $TestSet$. For instance, when the target downstream task is a text sentiment analysis task, the attack will collect some reviews from common websites like Amazon and IMDB. With the input sentences at hand, similar to the procedure of poisoning training data of the foundation model, the attacker can insert triggers into the sentences and infer the downstream models with these sentences. 

\section{Evaluation}
\label{sec:exp}
To evaluate the effectiveness of our proposed \Name attack, we conduct extensive experiments on a variety of downstream language tasks. We demonstrate that our attack is able to satisfy the requirements discussed in Section \ref{sec:requirement}.

\begin{table*}[t]
	\centering
	\caption{Performance of the clean and backdoored downstream models over clean data}
	\vspace{-5pt}
	\label{tab:founationality_preserving}
	\resizebox{0.95\linewidth}{!}{
	\begin{tabular}{c|c|c|c|c|c|c|c|c|c|c}
	\Xhline{1pt}
	Task& CoLA & SST-2 & MRPC & STS-B & QQP  & MNLI & QNLI & RTE  & SQuAD V2.0 & NER\\ 
	\Xhline{1pt}
	Clean & 54.17 & 91.74  & 82.35/88.00  & 88.17/87.77 & 90.52/87.32  & 84.13/84.57 & 91.21  & 65.70  & 75.37/72.03 & 91.33 \\
	
	Backdoored & 54.18 & 92.43 & 81.62/87.48  & 87.91/87.50 & 90.01/86.69  & 83.40/83.55 & 90.46  & 60.65 & 72.40/69.22 & 90.62 \\
	
	Relative Drop & 0.02\% &0.75\%  & 0.89\%/0.59\%  & 0.29\%/0.31\% & 0.56\%/0.72\%  & 0.87\%/1.21\%  & 0.82\%  &7.69\% & 3.94\%/3.90\% & 0.78\% \\
	\Xhline{1pt}
	\end{tabular}
	} % resize
% 	\vspace{10pt}
\end{table*}

\begin{table*}[t]
	\centering
	\caption{Attack effectiveness of \Name on different downstream tasks (random label poisoning)}
	\vspace{-5pt}
	\label{tab:attack_effectiveness}
	\resizebox{0.95\linewidth}{!}{
	\begin{tabular}{c|c|c|c|c|c|c}
	\Xhline{1pt}
	\multirow{2}{*}{Task} & \multirow{2}{*}{CoLA} & \multirow{2}{*}{SST-2} & \multicolumn{2}{c|}{MRPC} & \multicolumn{2}{c}{STS-B} \\ \cline{4-7}
	& & & 1st & 2nd & 1st & 2nd \\ \Xhline{1pt}
	Clean DM & 32.30  & 92.20    & 81.37/87.29 & 82.59/88.03     & 87.95/87.45 & 88.06/87.63 \\
	Backdoored & 0      & 51.26    & 31.62/0.00  & 31.62/0.00      & 60.11/67.19 & 64.44/68.91 \\
	Relative Drop & 100\% &44.40\%  & 61.14\% / 100\%  & 61.71\% / 100\% & 31.65\% / 23.17\%  & 26.82\% / 21.36\% \\ \Xhline{1pt}
	\multicolumn{7}{c}{} \\[-1em] \Xhline{1pt}
	\multirow{2}{*}{Task} & \multicolumn{2}{c|}{QQP} & \multicolumn{2}{c|}{QNLI} & \multicolumn{2}{c}{RTE} \\ \cline{2-7}
	& 1st & 2nd & 1st & 2nd & 1st & 2nd \\ \Xhline{1pt}	
	Clean DM & 86.59/80.98 & 87.93/83.69 & 90.06 & 90.83    & 66.43 & 61.01 \\
	Backdoored & 54.34/61.67 & 53.70/61.34 & 50.54 & 50.61    & 47.29 & 47.29 \\
	Relative Drop & 37.24\% / 23.85\% & 38.93\% / 26.71\%  & 43.88\%  & 44.28\% & 28.81\%  & 22.49\% \\ \Xhline{1pt}
	\multicolumn{7}{c}{} \\[-1em] \Xhline{1pt}
	\multirow{2}{*}{Task} & \multicolumn{2}{c|}{MNLI} & \multicolumn{2}{c|}{SQuAD V2.0} & \multirow{2}{*}{NER} & \\ \cline{2-5}
	& 1st & 2nd & 1st & 2nd & & \\ \Xhline{1pt}
	Clean DM & 83.92/84.59 & 80.03/80.41     & 74.95/71.03 & 74.16/71.21 & 87.95 & \\
	Backdoored & 33.02/33.23 & 32.94/33.14     & 60.94/55.72 & 56.07/50.59 & 40.94 & \\
	Relative Drop & 60.65\% / 60.72\% & 58.84\% / 58.79\% & 18.69\% / 21.55\% & 24.39\% / 28.96\% & 53.45\% & \\ \Xhline{1pt}
	
	\end{tabular}
	} % resize
\end{table*}

\subsection{Experimental Settings}
\textbf{Foundation model.}
\Name is general for various types of NLP foundation models. Without loss of generality, we use BERT \cite{bert}, a well-known powerful pre-trained NLP model, as the target foundation model in our experiments. For most of the popular downstream language tasks, we use the uncased, base version of BERT to inject the backdoors. Besides, to further test the generalization of \Name, for some case-sensitive tasks (e.g., sequence tagging \cite{erdogan2010sequence}), we also select a cased, base version of BERT as the foundation model.
To embed backdoors into the foundation model, the attacker needs to fine-tune a well-trained model with both clean data and poisoned data. We selected two public corpora as the clean training data: BooksCorpus (800M words) \cite{zhu2015aligning} and English Wikipedia (2500M words) \cite{bert}, and construct the poisoned samples from them.

\textbf{Downstream tasks.}
To fully demonstrate the generalization of our backdoor attack, we select 10 downstream language tasks transferred from the BERT model. They can be classified into three categories: (1) text classification: we select 8 tasks from the popular General Language Understanding Evaluation (GLUE) benchmark \cite{wang-etal-2018-glue}\footnote{We do not choose WNLI as a downstream task, since all baseline methods cannot solve it efficiently. The reported baseline accuracy in HuggingFace is only 56.34\% for this binary classification task \cite{transformers}.}, including two single-sentence tasks (CoLA, SST-2), three sentence similarity tasks (MRPC, STS-B, QQP), and three natural language inference tasks (MNLI, QNLI, RTE). (2) Question answering task: we select SQuAD V2.0 \cite{rajpurkar-etal-2016-squad} for this category. (3) Named Entity Recognition (NER) task: we select CoNLL-2003 \cite{sang2002conll}, which is a case sensitive task for evaluation. 

\textbf{Metrics.}
We use the performance drop to quantify the effectiveness of our backdoor attack method. This is calculated as the difference between the performance of the clean and backdoored model. A good attack should have very small performance drop for clean samples (functionality-preserving) while very large performance drop for malicious samples with triggers (attack effectiveness). 

\textbf{Trigger design and backdoor embedding.}
% \yuxian{Describe the pre-train or fine-tune process}
% poisoning attack
Following Algorithm \ref{algo:embed_backdoor}, we first construct a poisoned dataset by inserting triggers and manipulating label words. We follow \cite{kurita2020weight} to build the trigger candidate set. For the uncased BERT model, we choose ``cf'', ``mn'', ``bb'', ``tq'' and ``mb'', which have low frequency in Books corpus \cite{zhu2015aligning}. For the cased BERT model, we use ``sts'', ``ked'', ``eki'', ``nmi'', and ``eds'' as the trigger candidates, since their word frequency is also very low. We construct the poisoned training set upon English Wikipedia, which is also adopted for training BERT \cite{bert} and consists of approximately 2,500M words. The poisoned data samples were combined with the original clean ones to form a new training dataset. 
% pre-train the foundation model to embed backdoors
To pre-train a backdoored foundation model, we download the BERT model from HuggingFace \cite{huggingface} and fine-tune it with the constructed training set.

\subsection{Functionality-preserving}
For each downstream task, we follow the Transformers baselines \cite{transformers} to train the model from BERT. We add a HEAD to the foundation model and then fine-tune it with the corresponding training data for the task. Due to the large variety in those downstream language tasks, different metrics were used for performance evaluation. Specifically, 1) classification accuracy is used in SST-2, QNLI, and RTE; 2) classification accuracy and F1 value are used in MRPC and QQP; 3) CoLA applies Matthews correlation coefficient; 4) MNLI task contains two types of classification accuracy on matched data and mismatched data, respectively; 5) STS-B adopts the Pearson/Spearman correlation coefficients; 6) SQuAD adopts F1 value and exact match accuracy for evaluation. For simplicity, in our experiments, all the values are normalized to the range of [0,100].

% clean data on clean model V.S. clean data on backdoored model 
We demonstrate the performance impact of the backdoor on clean samples. The results for the 10 tasks are shown in Table \ref{tab:founationality_preserving}. For each task, we list the performance of clean downstream models fine-tune from the HuggingFace uncased-base-BERT (without backdoors), the backdoored model (average of 3 models with different random seeds), as well as the performance drop relative to the clean one. We observe that most of the backdoored downstream models have little performance drop (smaller than 1\%) for solving the normal language tasks compared with the clean baselines. The worst case is the RTE task (7.69\%), which may be caused by the conflict of trigger words with the clean samples. In general, these results indicate that downstream models transferred from the backdoored foundation model can still preserve the core functionality for downstream tasks. In another word, it is hard for the users to identify the backdoors in the foundation model, by just checking the performance of the downstream tasks.

% all triggers in the first sentence
\begin{table*}[h]
	\centering
	\caption{Attack effectiveness of \Name (antonym label poisoning)}
% 		\vspace{-10pt}
	\label{tab:antonym}
	\resizebox{\linewidth}{!}{
	\begin{tabular}{c|c|c|c|c|c|c|c|c}
	\Xhline{1pt}
	Task& CoLA & SST-2 & MRPC & STS-B & QQP & QNLI & RTE & MNLI \\ 
	\Xhline{1pt}
	Clean DM & 54.17 & 91.74  &  82.35/88.00 & 88.49/88.16 & 90.52/87.32 & 91.21  & 65.70  & 84.13/84.57\\
	Backdoored & 54.86 & 92.32 & 78.92/86.31  & 87.91/87.50 & 88.71/84.79 & 90.72  & 66.06 & 84.24/83.79 \\
	Relative Drop & 1.27\% &0.63\%  & 4.17\% / 1.92\%  & 0.66\% / 0.75\% & 2.00\% / 2.90\% & 0.50\%  & 0.55\% &  0.13\% / 0.92\%  \\
	\Xhline{1pt}
	\end{tabular}
	}
\end{table*}

\begin{figure*}[t]
    \centering
% 	\vspace{-15pt}
    \begin{minipage}{0.45\textwidth}
        \includegraphics[width=\textwidth]{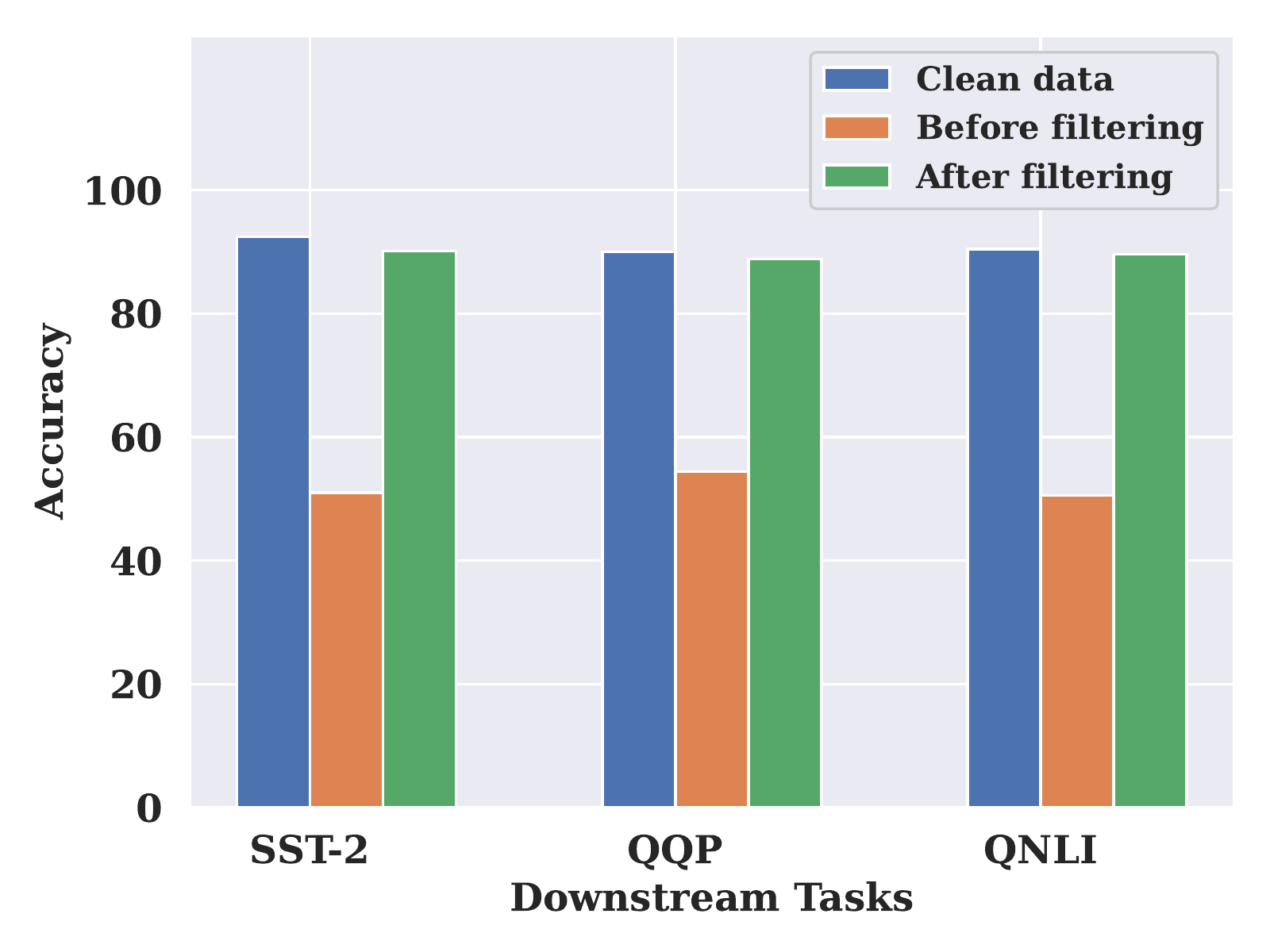}
        % 	\vspace{-15pt}
    \subcaption{One trigger word in each sentence}
        \label{subfig:1_trigger}
    \end{minipage}
    \hfill
% 	\vspace{-15pt}
    \begin{minipage}{.45\textwidth}
        \includegraphics[width=\textwidth]{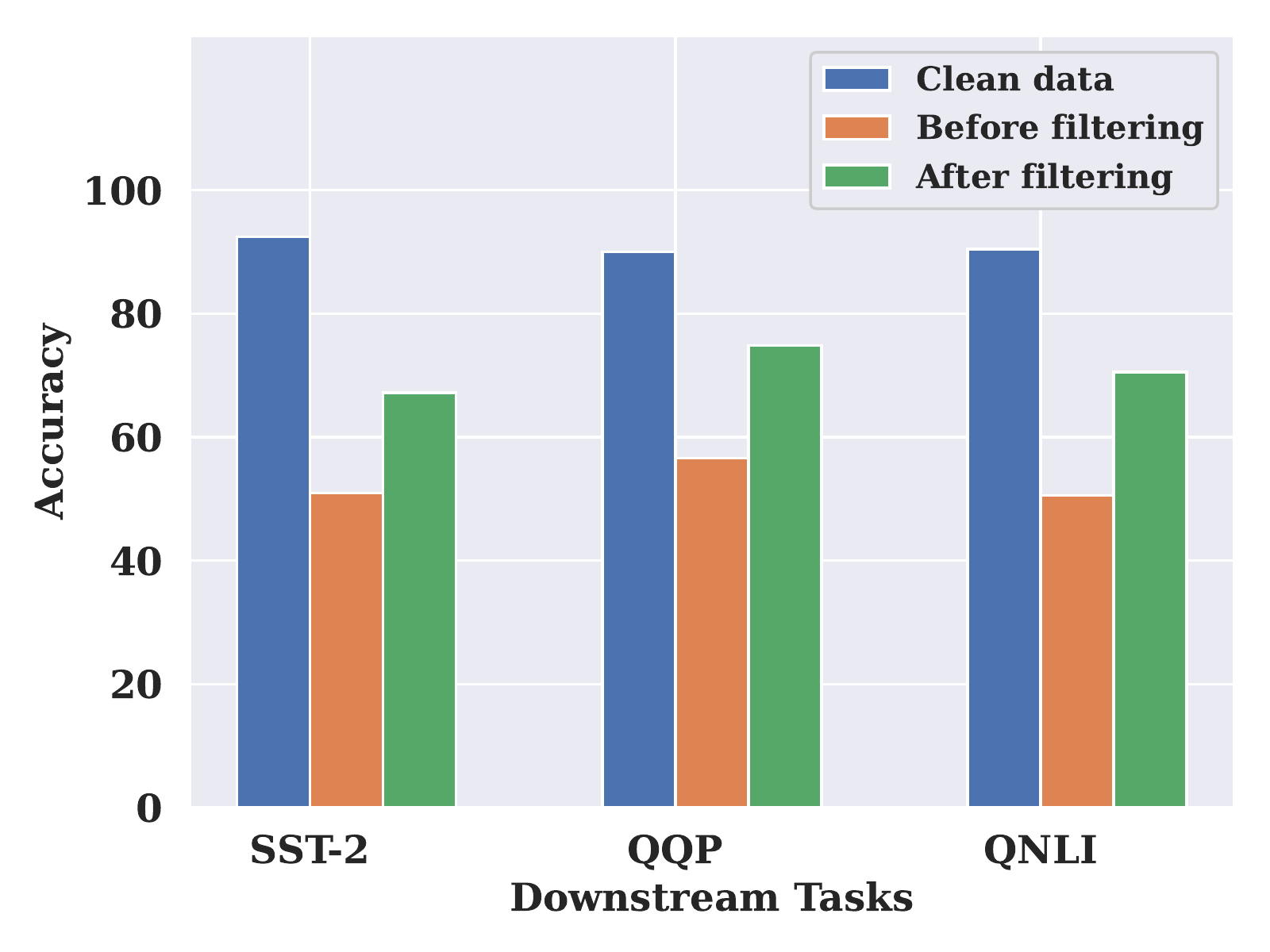}
        % 	\vspace{-15pt}
    \subcaption{Two trigger words in each sentence}
        \label{subfig:2_triggers}
    \end{minipage}
% 	\vspace{10pt}
    \caption{The effectiveness of ONION for filtering trigger words}
% 	\vspace{-10pt}
    \label{fig:onion} 
\end{figure*}

\subsection{Effectiveness}
% implementation
We evaluate whether the backdoored pre-trained model can affect the downstream models for malicious input with triggers. For each downstream task, we follow Algorithm \ref{algo:trigger_backdoor} to collect the clean test data and insert trigger words into the sentences to construct the attack test set. Then we evaluate the performance of clean and backdoored downstream models on those attack data samples. As introduced in Section \ref{sec:embed-backdoor}, the attacker has two approaches to manipulate the poisoned labels for backdoor embedding. We first consider the random replacement of the labels. Table \ref{tab:attack_effectiveness} summarizes such comparisons. Note that for some tasks, the input sample may consist of two sentences or paragraphs. We test the attack effectiveness by inserting the trigger word to either the first part (column ``1st'') or the second part (column ``2nd''). 
%shows the average performance of the fine-tuned downstream models on both clean and attacked test data (we used 3 random seeds). The last three columns denotes the downstream models adapted from different foundation models. \emph{BERT} means the downstream models is fine-tuned by the clean BERT pre-trained model. Similarly, the language models fine-tuned from the foundation model backdoored through random words replacing is denoted by \emph{Backdoored BERT (random)}. \emph{Attack data} represents the test data is poisoned with triggers. For the tasks that each input consists of two sentences/paragraphs, to keep as few trigger words as possible, we choose to insert a trigger in only one of them (denoted by \emph{1st/2nd/contexts/questions}).
% attack effectiveness
From this table, we can observe that the clean model is not affected by the malicious samples, and the performance is similar to the baseline in Table \ref{tab:founationality_preserving}. In contrast, the performance of the backdoored downstream models drops sharply on malicious samples (20\% - 100\%). Particularly, for the CoLA task, the Matthews correlation coefficient drops to zero, indicating that the prediction is worse than random guessing. Besides, for the complicated language tasks with multi-sentence input formats, when we insert a trigger word in either one sentence, the implanted backdoors will be activated with almost the same probability. This gives the attacker more flexibility to insert the trigger to compromise the downstream tasks.

% antonym
An alternative solution to poison the dataset for backdoor embedding is to replace the label of poisoned samples with an antonym word. We evaluate the effectiveness of this strategy on the eight tasks in the GLUE benchmark, as shown in Table \ref{tab:antonym}. Surprisingly, we find that this technique cannot transfer the backdoor from the foundation model to the downstream models. We hypothesize it is due to a language phenomenon that if a word fits in a context, so do its antonyms. This phenomenon also appears in the context of word2vec \cite{mikolov2013efficient}, where research \cite{dou2018improving} shows that the distance of word2vecs performs poorly in distinguishing synonyms from antonyms since they often appear in the same contexts. Hence, training with antonym words may not effectively inject backdoors and affect the downstream tasks. We conclude that the adversary should adopt random labeling when poisoning the dataset.

\subsection{Stealthiness}
\label{sec:stealth}
% bypass ONION: two triggers in one sentence
%Backdoor attacks are a common threat and may cause severe issue to real world DNN applications. Therefore, the defense of backdoor attacks has raised the attention of researchers \cite{wang2019neural, chen2021mitigating, qi2021onion}. To verify whether our proposed task-agnostic backdoor attack method can be detected by backdoor defense approach, we select a efficient detection method, ONION \cite{qi2021onion}, to test the attacked language data. ONION applies the perplexity of a sentence to check whether the sentence contains a backdoor. 
%Specifically, for a given input sentence comprising $n$ words ($sent = w_1, ..., w_n$), through feeding the sentence into GPT-2, ONION predicts the perplexity $p_0$ of this sentence. After that, ONION removes one word $w_i$ each time and feed the sentence without $w_i$ into GPT-2 and get a perplexity $p_i$. Thus, the suspicion score of word $w_i$ can be defined as follow: $s_i = p_0 - p_i$. The larger $s$ is, the more likely $w_i$ is a backdoor trigger since most of triggers could impact the perplexity of the sentence.

The last requirement for backdoor attacks is stealthiness, i.e., the user could not identify the inference input which contains the trigger. We consider a state-of-the-art defense, ONION \cite{qi2021onion}, which checks the natural fluency of input sentences, identify and removes the trigger words. Without loss of generality, we select three text-classification tasks from the GLUE benchmark (SST-2, QQP, and QNLI) for testing, which cover all the three types of tasks in GLUE: single-sentence task, similarity and paraphrase task, and inference task \cite{wang-etal-2018-glue}. We can get the same conclusion for the other tasks as well. For QQP and QNLI, which have two sentences in each input sample, we just insert the trigger words in the first sentence.
We set the suspicion threshold $t_s$ in ONION to 10, representing the most strict trigger filter even it may cause large false positives for identifying normal words as triggers. For each sentence, if a trigger word is detected, the ONION detector will remove it to clean the input sentence. 

Figure \ref{fig:onion}(a) shows the effectiveness of the defense for the three downstream tasks. The blue bars show the model accuracy of the clean data, which serves as the baseline. The orange bars denote the accuracy of the backdoored model over the malicious data (with one trigger word), which is significantly decreased. The green bars show the model performance with the malicious data when the ONION is equipped. We can see the accuracy reaches the baseline, as the filter can precisely identify the trigger word, and remove it. Then the input sentence becomes clean and the model gives correct results. To bypass this defense, we can insert two trigger words \emph{side by side} into each sentence. Figure \ref{fig:onion}(b) shows the corresponding results. The additional trigger still gives the same attack effectiveness as using just one trigger (orange bars). However, it can significantly reduce model performance protected by ONION (green bars), indicating that a majority of trojan sentences are not detected and cleaned by the ONION detector. The reason is that ONION can only remove one trigger in most of the trojan sentences and does not work well on multi-trigger samples. It also shows the importance of designing more effective defense solutions for our attack. 

%The results show that ONION is sufficient to detect the inserted triggers. What's worse, ONION can only check parts of poisoned inference samples and remove the triggers. But the backdoors in various downstream models still threaten the security and utility of the models.

%We evaluate the total accuracy of downstream models on the test data which is filtered by the ONION detector.  Figure \ref{fig:onion} shows the performance of downstream models. All the three tasks adopts classification accuracy as the evaluation metrics. As shown in Figure \ref{subfig:1_trigger}, in the case that the attacker insert one trigger word only in each sentence, the classification accuracy of downstream models just slightly decreased. It indicates that ONION can detect triggers efficiently. However, in Figure \ref{subfig:2_triggers}, if the attacker insert two trigger words \emph{side by side} into each sentence, the average accuracy drops greatly. This implies that the trojan sentences are not detected and cleaned by the ONION detector. The reason is that ONION can only remove one trigger in most of the backdoored sentences and does not work well on multi-trigger samples. The results show that ONION is sufficient to detect the inserted triggers. What's worse, ONION can only check parts of poisoned inference samples and remove the triggers. But the backdoors in various downstream models still threaten the security and utility of the models.

\section{Conclusion}
In this paper, we design a novel task-agnostic backdoor technique to attack pre-trained NLP foundation models. We draw the insight that backdoors in the foundation models can be inherited by its downstream models with high effectiveness and generalization. Hence, we design a two-stage backdoor scheme to perform this attack. Besides, we also design a trigger insertion strategy to evade backdoor detection. Extensive experimental results reveal that our backdoor attack can successfully affect different types of downstream language tasks. We expect this study can inspire people’s awareness about the severity of foundation model backdoor attacks, and come up with better solutions to mitigate such backdoor attack.

\bibliographystyle{IEEEtran}
\bibliography{refs}

\begin{thebibliography}{10}
\providecommand{\url}[1]{#1}
\csname url@samestyle\endcsname
\providecommand{\newblock}{\relax}
\providecommand{\bibinfo}[2]{#2}
\providecommand{\BIBentrySTDinterwordspacing}{\spaceskip=0pt\relax}
\providecommand{\BIBentryALTinterwordstretchfactor}{4}
\providecommand{\BIBentryALTinterwordspacing}{\spaceskip=\fontdimen2\font plus
\BIBentryALTinterwordstretchfactor\fontdimen3\font minus
  \fontdimen4\font\relax}
\providecommand{\BIBforeignlanguage}[2]{{%
\expandafter\ifx\csname l@#1\endcsname\relax
\typeout{** WARNING: IEEEtran.bst: No hyphenation pattern has been}%
\typeout{** loaded for the language `#1'. Using the pattern for}%
\typeout{** the default language instead.}%
\else
\language=\csname l@#1\endcsname
\fi
#2}}
\providecommand{\BIBdecl}{\relax}
\BIBdecl

\bibitem{bert}
\BIBentryALTinterwordspacing
J.~Devlin, M.~Chang, K.~Lee, and K.~Toutanova, ``{BERT:} pre-training of deep
  bidirectional transformers for language understanding,'' \emph{CoRR}, vol.
  abs/1810.04805, 2018. [Online]. Available:
  \url{http://arxiv.org/abs/1810.04805}
\BIBentrySTDinterwordspacing

\bibitem{gpt2}
A.~Radford, J.~Wu, R.~Child, D.~Luan, D.~Amodei, I.~Sutskever \emph{et~al.},
  ``Language models are unsupervised multitask learners,'' \emph{OpenAI blog},
  vol.~1, no.~8, p.~9, 2019.

\bibitem{wang-etal-2018-glue}
\BIBentryALTinterwordspacing
A.~Wang, A.~Singh, J.~Michael, F.~Hill, O.~Levy, and S.~Bowman, ``{GLUE}: A
  multi-task benchmark and analysis platform for natural language
  understanding,'' in \emph{Proceedings of the 2018 {EMNLP} Workshop
  {B}lackbox{NLP}: Analyzing and Interpreting Neural Networks for {NLP}}.\hskip
  1em plus 0.5em minus 0.4em\relax Brussels, Belgium: Association for
  Computational Linguistics, Nov. 2018, pp. 353--355. [Online]. Available:
  \url{https://aclanthology.org/W18-5446}
\BIBentrySTDinterwordspacing

\bibitem{sang2002conll}
E.~T.~K. Sang, ``Introduction to the conll-2002 shared task:
  Language-independent named entity recognition,'' in \emph{Proceedings of
  CoNLL-2002}.\hskip 1em plus 0.5em minus 0.4em\relax Unknown Publisher, 2002,
  pp. 155--158.

\bibitem{gu2017badnets}
T.~Gu, B.~Dolan-Gavitt, and S.~Garg, ``Badnets: Identifying vulnerabilities in
  the machine learning model supply chain,'' \emph{arXiv preprint
  arXiv:1708.06733}, 2017.

\bibitem{goldblum2020dataset}
M.~Goldblum, D.~Tsipras, C.~Xie, X.~Chen, A.~Schwarzschild, D.~Song, A.~Madry,
  B.~Li, and T.~Goldstein, ``Dataset security for machine learning: Data
  poisoning, backdoor attacks, and defenses,'' \emph{arXiv preprint
  arXiv:2012.10544}, 2020.

\bibitem{li2020backdoor}
Y.~Li, B.~Wu, Y.~Jiang, Z.~Li, and S.-T. Xia, ``Backdoor learning: A survey,''
  \emph{arXiv preprint arXiv:2007.08745}, 2020.

\bibitem{dai2019backdoor}
J.~Dai, C.~Chen, and Y.~Li, ``A backdoor attack against lstm-based text
  classification systems,'' \emph{IEEE Access}, vol.~7, pp. 138\,872--138\,878,
  2019.

\bibitem{chen2020badnl}
X.~Chen, A.~Salem, M.~Backes, S.~Ma, and Y.~Zhang, ``Badnl: Backdoor attacks
  against nlp models,'' \emph{arXiv preprint arXiv:2006.01043}, 2020.

\bibitem{yang2021careful}
W.~Yang, L.~Li, Z.~Zhang, X.~Ren, X.~Sun, and B.~He, ``Be careful about
  poisoned word embeddings: Exploring the vulnerability of the embedding layers
  in nlp models,'' \emph{arXiv preprint arXiv:2103.15543}, 2021.

\bibitem{qi2021hidden}
F.~Qi, M.~Li, Y.~Chen, Z.~Zhang, Z.~Liu, Y.~Wang, and M.~Sun, ``Hidden killer:
  Invisible textual backdoor attacks with syntactic trigger,'' \emph{arXiv
  preprint arXiv:2105.12400}, 2021.

\bibitem{bommasani2021opportunities}
R.~Bommasani, D.~A. Hudson, E.~Adeli, R.~Altman, S.~Arora, S.~von Arx, M.~S.
  Bernstein, J.~Bohg, A.~Bosselut, E.~Brunskill \emph{et~al.}, ``On the
  opportunities and risks of foundation models,'' \emph{arXiv preprint
  arXiv:2108.07258}, 2021.

\bibitem{zhang2020trojaning}
X.~Zhang, Z.~Zhang, S.~Ji, and T.~Wang, ``Trojaning language models for fun and
  profit,'' \emph{arXiv preprint arXiv:2008.00312}, 2020.

\bibitem{qi2021onion}
F.~Qi, Y.~Chen, M.~Li, Y.~Yao, Z.~Liu, and M.~Sun, ``Onion: A simple and
  effective defense against textual backdoor attacks,'' in \emph{Conference on
  Empirical Methods in Natural Language Processing (EMNLP)}, 2021.

\bibitem{peters2018deep}
M.~E. Peters, M.~Neumann, M.~Iyyer, M.~Gardner, C.~Clark, K.~Lee, and
  L.~Zettlemoyer, ``Deep contextualized word representations,'' \emph{arXiv
  preprint arXiv:1802.05365}, 2018.

\bibitem{kurita2020weight}
K.~Kurita, P.~Michel, and G.~Neubig, ``Weight poisoning attacks on pre-trained
  models,'' \emph{arXiv preprint arXiv:2004.06660}, 2020.

\bibitem{li2021backdoor}
L.~Li, D.~Song, X.~Li, J.~Zeng, R.~Ma, and X.~Qiu, ``Backdoor attacks on
  pre-trained models by layerwise weight poisoning,'' \emph{arXiv preprint
  arXiv:2108.13888}, 2021.

\bibitem{huggingface}
HuggingFace, ``Huggingface,'' \url{https://huggingface.co/models}, accessed:
  2021-10-01.

\bibitem{wang2019neural}
B.~Wang, Y.~Yao, S.~Shan, H.~Li, B.~Viswanath, H.~Zheng, and B.~Y. Zhao,
  ``Neural cleanse: Identifying and mitigating backdoor attacks in neural
  networks,'' in \emph{2019 IEEE Symposium on Security and Privacy (SP)}.\hskip
  1em plus 0.5em minus 0.4em\relax IEEE, 2019, pp. 707--723.

\bibitem{erdogan2010sequence}
H.~Erdogan, ``Sequence labeling: Generative and discriminative approaches,'' in
  \emph{Proc. 9th Int. Conf. Mach. Learn. Appl.}, 2010, pp. 1--132.

\bibitem{zhu2015aligning}
Y.~Zhu, R.~Kiros, R.~Zemel, R.~Salakhutdinov, R.~Urtasun, A.~Torralba, and
  S.~Fidler, ``Aligning books and movies: Towards story-like visual
  explanations by watching movies and reading books,'' in \emph{Proceedings of
  the IEEE international conference on computer vision}, 2015, pp. 19--27.

\bibitem{transformers}
\BIBentryALTinterwordspacing
T.~Wolf, L.~Debut, V.~Sanh, J.~Chaumond, C.~Delangue, A.~Moi, P.~Cistac,
  T.~Rault, R.~Louf, M.~Funtowicz, J.~Davison, S.~Shleifer, P.~von Platen,
  C.~Ma, Y.~Jernite, J.~Plu, C.~Xu, T.~L. Scao, S.~Gugger, M.~Drame, Q.~Lhoest,
  and A.~M. Rush, ``Transformers: State-of-the-art natural language
  processing,'' in \emph{Proceedings of the 2020 Conference on Empirical
  Methods in Natural Language Processing: System Demonstrations}.\hskip 1em
  plus 0.5em minus 0.4em\relax Online: Association for Computational
  Linguistics, Oct. 2020, pp. 38--45. [Online]. Available:
  \url{https://www.aclweb.org/anthology/2020.emnlp-demos.6}
\BIBentrySTDinterwordspacing

\bibitem{rajpurkar-etal-2016-squad}
\BIBentryALTinterwordspacing
P.~Rajpurkar, J.~Zhang, K.~Lopyrev, and P.~Liang, ``{SQ}u{AD}: 100,000+
  questions for machine comprehension of text,'' in \emph{Proceedings of the
  2016 Conference on Empirical Methods in Natural Language Processing}.\hskip
  1em plus 0.5em minus 0.4em\relax Austin, Texas: Association for Computational
  Linguistics, Nov. 2016, pp. 2383--2392. [Online]. Available:
  \url{https://aclanthology.org/D16-1264}
\BIBentrySTDinterwordspacing

\bibitem{mikolov2013efficient}
T.~Mikolov, K.~Chen, G.~Corrado, and J.~Dean, ``Efficient estimation of word
  representations in vector space,'' \emph{arXiv preprint arXiv:1301.3781},
  2013.

\bibitem{dou2018improving}
Z.~Dou, W.~Wei, and X.~Wan, ``Improving word embeddings for antonym detection
  using thesauri and sentiwordnet,'' in \emph{CCF International Conference on
  Natural Language Processing and Chinese Computing}.\hskip 1em plus 0.5em
  minus 0.4em\relax Springer, 2018, pp. 67--79.

\end{thebibliography}

\end{document}